\documentclass[journal]{IEEEtran}

\usepackage{hyperref}
\usepackage{graphicx}
\usepackage{multirow}
\usepackage{cite}
\usepackage{amsmath}
\usepackage{amssymb}
\usepackage{color}
\ifCLASSOPTIONcompsoc
  \usepackage[caption=false,font=normalsize,labelfont=sf,textfont=sf]{subfig}
\else
  \usepackage[caption=false,font=footnotesize]{subfig}
\fi

\begin{document}

\title{A Strong Baseline and Batch Normalization Neck for Deep Person Re-identification}

\author{Hao Luo,
        Wei Jiang,
        Youzhi Gu,
        Fuxu Liu,
        Xingyu Liao,
        Shenqi Lai,
        Jianyang Gu
\thanks{This paper is the extended version of the oral paper \cite{luo2019bags} accepted by TRMTMCT2019 Workshop in CVPR2019.}
\thanks{Hao Luo, Wei Jiang, Youzhi Gu, Jianyang Gu is with the State Key Laboratory of Industrial Control Technology, College of Control Science and Enginneering, Zhejiang University, Hangzhou 310027, China; E-mail: haoluocsc@zju.edu.cn, jiangwei\_zju@zju.edu.cn.}
\thanks{Fuxu Liu is with Ping An Technology, Shenzhen, China; E-mail: LIUFUXU641@pingan.com.cn}
\thanks{Xingyu Liao is with Chinese Academy of Sciences, Beijing , China; E-mail: randall@mail.ustc.edu.cn}
\thanks{Shenqi Lai is with Xi'an Jiaotong University, Xi'an, China; E-mail: laishenqi@stu.xjtu.edu.cn}
\thanks{Manuscript received June 19, 2019; revised MMDD, YYYY.}}

\markboth{Journal of \LaTeX\ Class Files,~Vol.~14, No.~8, August~2015}%
{Shell \MakeLowercase{\textit{et al.}}: Bare Demo of IEEEtran.cls for IEEE Journals}
\maketitle

\begin{abstract}
This study proposes a simple but strong baseline for deep person re-identification (ReID).
Deep person ReID has achieved great progress and high performance in recent years.
However, many state-of-the-art methods design complex network structures and concatenate multi-branch features.
In the literature, some effective training tricks briefly appear in several papers or source codes.
The present study collects and evaluates these effective training tricks in person ReID.
By combining these tricks, the model achieves 94.5\% rank-1 and 85.9\% mean average precision on Market1501 with only using the global features of ResNet50.
The performance surpasses all existing global- and part-based baselines in person ReID. We propose a novel neck structure named as batch normalization neck (BNNeck).
BNNeck adds a batch normalization layer after global pooling layer to separate metric and classification losses into two different feature spaces because we observe they are inconsistent in one embedding space.
Extended experiments show that BNNeck can boost the baseline, and our baseline can improve the performance of existing state-of-the-art methods.
Our codes and models are available at: \href{https://github.com/michuanhaohao/reid-strong-baseline}{https://github.com/michuanhaohao/reid-strong-baseline}
\end{abstract}


\IEEEpeerreviewmaketitle

\section{Introduction}
Person re-identification (ReID) is widely applied in video surveillance and criminal investigation applications \cite{wang2019incremental}.
Person ReID with deep neural networks has progressed and achieved high performance in recent years \cite{LUO2019,wang2018mancs,sun2018beyond}.
Apart from many novel and effective ideas being proposed, the improvement of baseline model plays a key role.
For example, Jian \emph{et al.} \cite{jian2019extended,jian2018integrating} proposed a baseline that greatly promoted development of underwater saliency detection.
The importance of baseline model should not be ignored. However, few works \cite{zheng2018discriminatively,xiong2019good,sun2018beyond} have focused on the design of an effective baseline. The performance of such baselines has gradually become obsolete due to the rapid development of person ReID. In the literature, some effective training tricks or refinements briefly appear in several papers or source codes. In the present study, we design a strong and effective baseline for person ReID by collecting and evaluating such effective training tricks.


\begin{figure}[htb]
\centering
\includegraphics[width=0.45\textwidth]{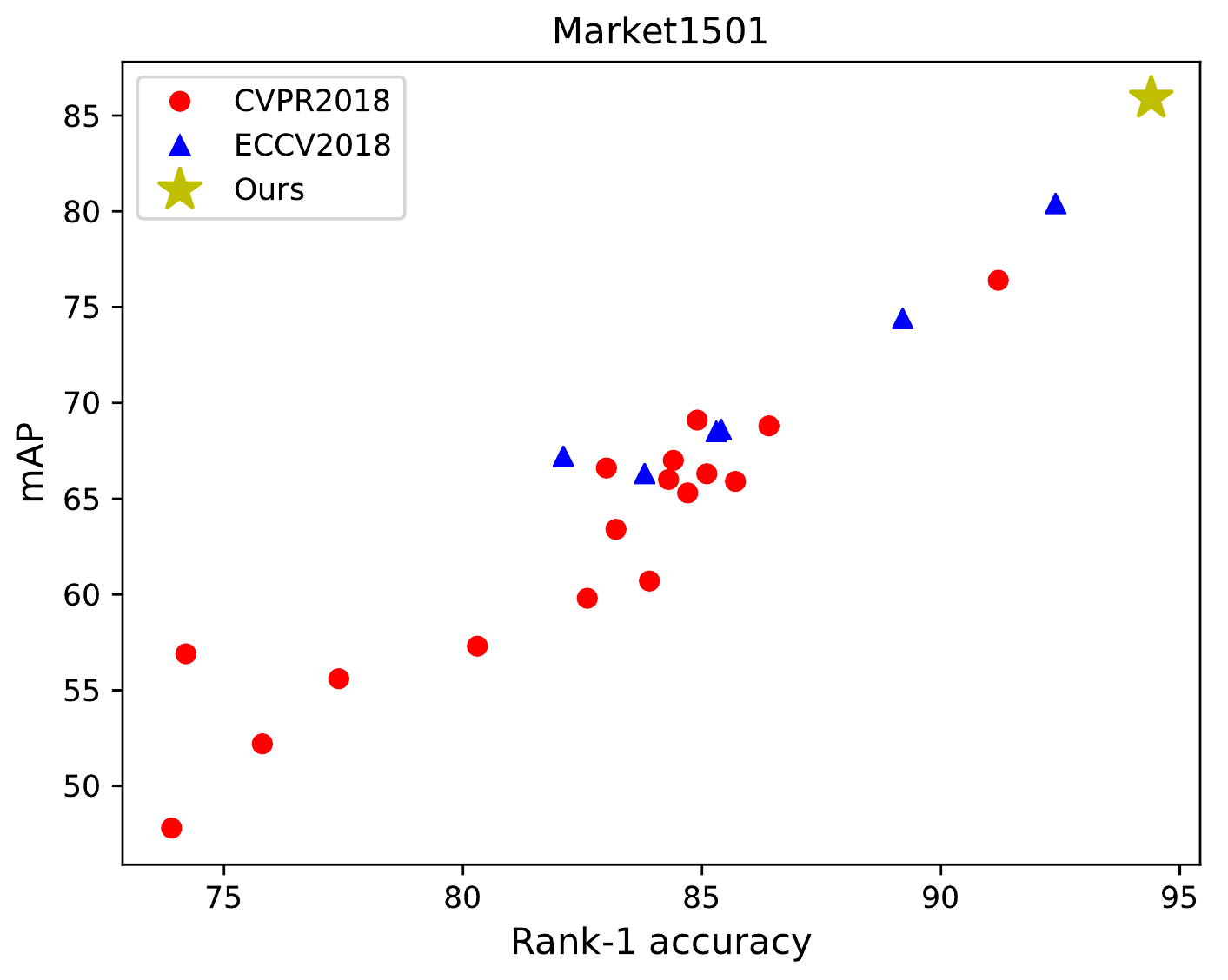}
\caption{Performance of different baselines on Market1501. Our strong baseline are compared with other baselines published on ECCV2018 and CVPR2018.}
\label{fig:baseline}
\end{figure}

This study has three motivations. First, we survey papers published on ECCV and CVPR in 2018.
As shown in Fig. \ref{fig:baseline}, many previous works were expanded on poor baselines. Only two in 23 baselines surpassed 90\% rank-1 accuracy on Market1501.
The rank-1 accuracies of four baselines were even lower than 80\%.
Achieving improvements on poor baselines cannot strictly demonstrate the effectiveness of some methods. Thus, a strong baseline is crucial in promoting research development.

Second, we discover that the improvements of some works were mainly from training tricks rather than methods themselves.
So they were unfairly compared with other state-of-the-art methods.
However, the training tricks were understated in the paper; thus, readers ignored them, thereby exaggerating the effectiveness of the method. We suggest that reviewers consider these tricks when commenting on academic papers.

Third, the industry prefers simple but effective models over concatenating many local features in the inference stage.
In order to achieve great performance, researchers in the academia always combine several local features or use semantic information from pose estimation or segmentation models. Nevertheless, such methods bring extra consumption. Large features also greatly reduce the speed of the retrieval process. Thus, we use tricks to improve the capability of the ReID model and only use global features extracted by the model.

On the basis of the aforementioned considerations, the motivations of designing a strong baseline are summarized as follows:
\begin{itemize}
  \item For the academia, we survey many works published on top conferences and discover that most of them were expanded on poor baselines. We aim to provide a strong baseline for researchers to achieve high accuracies in person ReID.
  \item For the community, we aim to provide references to reviewers regarding tricks that will affect the performance of the ReID model. We suggest that reviewers consider these tricks when comparing the performance of different methods.
  \item For the industry, we aim to provide effective tricks for acquiring improved models without extra consumption.
\end{itemize}

Many effective training tricks have been presented in papers or open-sourced projects. We collect tricks and evaluate each of them on ReID datasets. After numerous experiments, we select six tricks to introduce in this study. We propose a novel bottleneck structure, namely, batch normalization neck (BNNeck). As classification and metric losses are inconsistent in the same embedding space, BNNeck optimizes these two losses in two different embedding spaces. In addition, person ReID task mainly focuses on ranking performance, such as cumulative match characteristic (CMC) curve and mAP, but ignores the clustering effect, such as intra-class compactness and inter-class separability. However, clustering effect is important to some special tasks, such as object tracking, which must decide on a distance threshold to separate positive samples from negative ones. An easy approach to overcome this problem is to train the model with center loss. Finally, we add the tricks into a widely used baseline to obtain our modified baseline (the backbone is ResNet50), which achieves 94.5\% and 85.9\% mAP on Market1501.

To determine whether these tricks are generally useful or not, we design extended experiments from three aspects.
First, we follow the cross-domain ReID settings in which the models are trained and evaluated on different datasets.
Cross-domain experiments can show whether the tricks boost the models or simply suppress overfitting in the training dataset.
Second, we evaluate all tricks with different backbones, such as ResNet18, SeResNet50, and IBNNet-50.
All backbones achieve improvements from our training tricks.
Third, we reproduce some state-of-the-art methods on our modified baseline.
Experimental results show that our baseline obtains better performance than those reported in published papers.
Although our baseline achieves surprising performance, some methods remain effective on our baseline.
Thus, our baseline can be a strong baseline for the ReID community.

There are four main differences between the original version and this paper. 1)We add a new section Related Works to introduce the developments of deep person ReID.
2) We perform a lot of extended experiments to explain the effectiveness of BNNeck.
3) We explain why we introduce center loss and perform some extended experiments to analyze the effect of center loss.
4) We evaluate different backbones and some state-of-the-art methods on our baseline.

The contributions of this study are summarized as follows:
\begin{itemize}
  \item We collect effective training tricks for person ReID. The improvements from each trick are evaluated on two widely used datasets.
  \item We observe the inconsistency between ID loss and triplet and propose a novel neck structure, namely, BNNeck.
  \item We observe that the ReID task ignores intra-class compactness and inter-class separability and claim that center loss can compensate for it.
  \item We proposed a strong ReID baseline. With ResNet50 backbone, it achieves 94.5\% and 85.9\% mAP on Market1501. To our best knowledge, this result is the best performance acquired by global features in person ReID.
  \item We design extended experiments to demonstrate that our baseline can be a strong baseline for the ReID community.
\end{itemize}

\section{Related Works}
This section focuses on deep learning baseline for person ReID. In addition, existing approaches compared with our strong baseline for deep person ReID are introduced.

\subsection{Baseline for Deep Person ReID}
Recent studies on person ReID mostly focus on building deep convolutional neural networks (CNNs) to represent the features of person images in an end-to-end learning manner.
GoogleNet \cite{szegedy2015going}, ResNet \cite{he2016deep}, DenseNet \cite{huang2017densely}, etc are widely used backbone networks.
The baselines can be classified into two main genres in accordance with the loss function, \emph{i.e.} classification loss and metric loss.
For classification loss, Zheng \emph{et al.} \cite{zheng2018discriminatively} proposed ID-discriminative embedding (IDE) to train the re-ID model as image classification which is fine-tuned from the ImageNet \cite{deng2009imagenet} pre-trained models.
Classification loss is also called ID loss in person ReID because IDE is trained by classification loss. However, ID loss requires an extra fully connected (FC) layer to predict the logits of person IDs in the training stage. In the inference stage, such FC layer is removed, and the feature from the last pooling layer is used as the representation vector of the person image.

Different from ID loss, metric loss regards the ReID task as a clustering or ranking problem. The most widely used baseline based on metric learning is training model with triplet loss \cite{liu2017end}.
A triplet includes there images, \emph{i.e.} anchor, positive, and negative samples.
The anchor and positive samples belong to the same person ID, whereas the negative sample belongs to a different person ID. Triplet loss minimizes the distance from the anchor sample to the positive sample and maximizes the distance from the anchor sample to the negative one.
However, triplet loss is greatly influenced by the sample triplets.
Inspired by FaceNet\cite{schroff2015facenet}, Hermans \emph{et al.} proposed an online hard example mining for triplet loss (TriHard loss).
Most current methods are expanded on the TriHard baseline. Combining ID loss with TriHard loss is also a popular manner of acquiring a strong baseline \cite{LUO2019}.

Apart from designing different losses, some works focus on building effective baseline model for deep person ReID.
In \cite{xiong2019good}, three good practices were proposed to build an effective CNN baseline toward person ReID. Their most important practice is adding a batch normalization (BN) layer after the global pooling layer. Similar to these models, the baseline uses a global feature for image representation.
Sun \emph{et al.} \cite{sun2018beyond} proposed part-based convolutional baseline (PCB). Given an image input, PCB outputs a convolutional descriptor consisting of several part-level features. Both baselines have achieved good performance in person ReID.

\subsection{Some Existing Approaches for Deep person ReID}
On the basis of the aforementioned baselines, many methods have been proposed in the past few years. We divide these works into striped-based, pose-guided, mask-guided, attention-based, GAN-based, and re-ranking methods.

\textbf{Stripe-based methods}, which divide the image into several stripes and extract local features for each stripe, play an important role in person ReID.
Inspired by PCB, the typical methods includes AlignedReID++\cite{LUO2019}, MGN \cite{wang2018learning}, SCPNet \cite{fan2018scpnet}, etc.
Stripe-based local features are effective in boosting the performance of the ReID model. However, they always encounter the problem of pose misalignment.

\textbf{Pose-guided methods} \cite{zhao2017spindle, wei2017glad, Sarfraz_2018_CVPR, zheng2019pose} use an extra pose/skeleton estimation model to acquire human pose information.
Pose information can exactly align corresponding parts of two person images. However, an extra model brings additional computation consumption. A trade off between the performance and speed of the model is important.

\textbf{Mask-guided models} \cite{song2018mask,kalayeh2018human,qi2018maskreid} use mask as external cues to remove the background clutters in pixel level and contain body shape information.
For example, Song \emph{et al.} \cite{song2018mask} proposed a mask-guided contrastive attention model that applies binary segmentation masks to learn features separately from the body and background regions.
Kalayeh \emph{et al.} \cite{kalayeh2018human} proposed SPReID, which uses human semantic parsing to harness local visual cues for person ReID. Mask-guided models extremely rely on accurate pedestrian segmentation model.

\textbf{Attention-based methods} \cite{si2018dual,li2018harmonious,li2018diversity,xu2018attention} involve an attention mechanism to extract additional discriminative features. In comparison with pixel-level masks, attention region can be regraded as an automatically learned high-level `mask'.
A popular model is Harmonious Attention CNN (HA-CNN) model proposed by Li \emph{et al.} \cite{li2018harmonious}.
HA-CNN combines the learning of soft pixel and hard regional attentions along with simultaneous optimization of feature representations.
An advantage of attention-based models is that they do not require a segmentation model to acquire mask information.

\textbf{GAN-based methods} \cite{zheng2017unlabeled,wei2018person,zhong2019camstyle,qian2018pose,zheng2019joint,zhao2018adversarial} address the limited data for person ReID.
Zheng \emph{et al.} \cite{zheng2017unlabeled} first used GAN \cite{goodfellow2014generative} to generate images for enriching ReID datasets.
The GAN model randomly generates unlabeled and unclear images.
On the basis of \cite{zheng2017unlabeled}, PTGAN \cite{wei2018person} and CamStyle \cite{zhong2019camstyle} were proposed to bridge domain and camera gaps for person ReID, respectively.
Qian \emph{et al.} \cite{qian2018pose} proposed PNGAN for obtaining a new pedestrian feature and transforming a person into normalized poses. The final feature is obtained by combining the pose-independent features with original ReID features. With the development of GAN, many ganbased methods have been proposed to generate high quality for supervised and unsupervised person ReID tasks.
Zhao \emph{et al.} \cite{zhao2018adversarial} integrated GAN with a widely used triplet loss method.

\textbf{Re-ranking methods} \cite{zhong2017re,shen2018deep,ye2016person,bai2019re} are post-processing strategies for image retrieval.
In general, person ReID simply uses Euclidean or cosine distances in the retrieval stage.
Zhong \emph{et al.} \cite{zhong2017re} a $k$-reciprocal encoding method with the Jaccard distance of probe and gallery images to re-rank the ReID results.
Shen \emph{et al.} \cite{shen2018deep} proposed a deep group-shuffling random walk (DGRW) network for fully utilizing the affinity information between gallery images in training and testing processes. In the retrieval stage, DGRW can be regarded as a re-ranking method. Re-ranking is a critical step in improving retrieval accuracy.

\begin{figure*}[htt]
\centering
\subfloat[The pipeline of the standard baseline.]{
\begin{minipage}[b]{0.7\textwidth}
\includegraphics[width=1\textwidth]{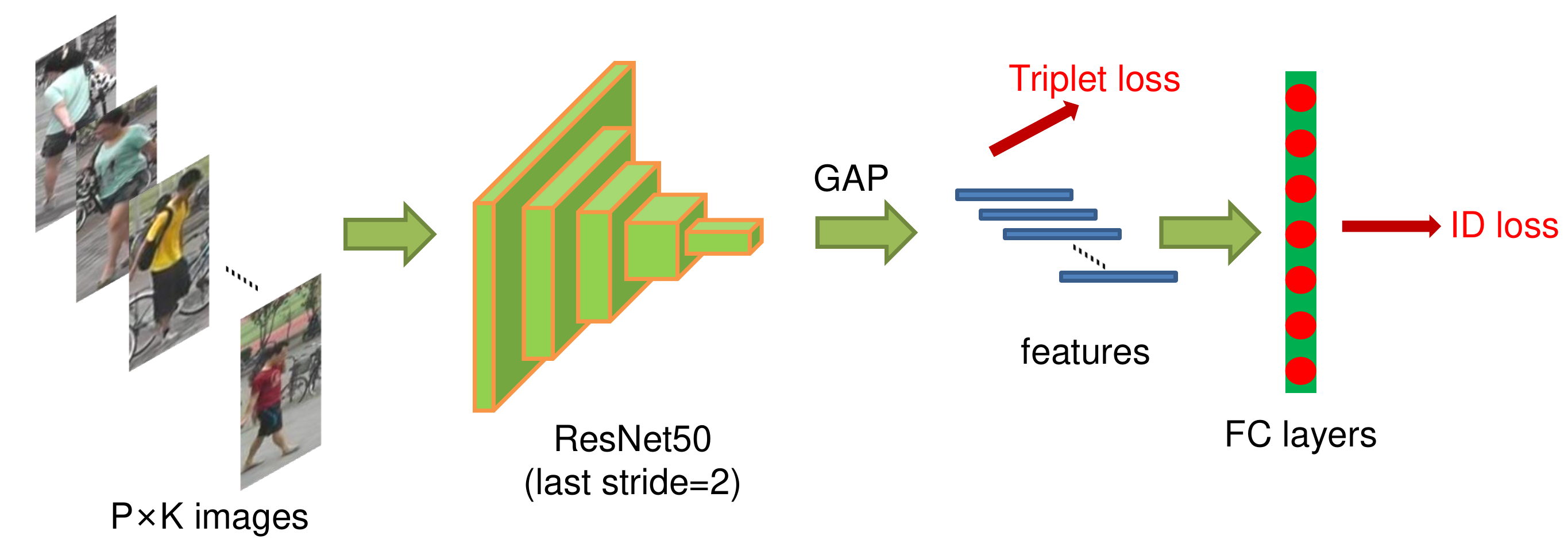}
\end{minipage}
\label{fig:arc_fisrt}}
\hfil
\subfloat[The pipeline of our modified baseline.]{
\begin{minipage}[b]{0.9\textwidth}
\includegraphics[width=1\textwidth]{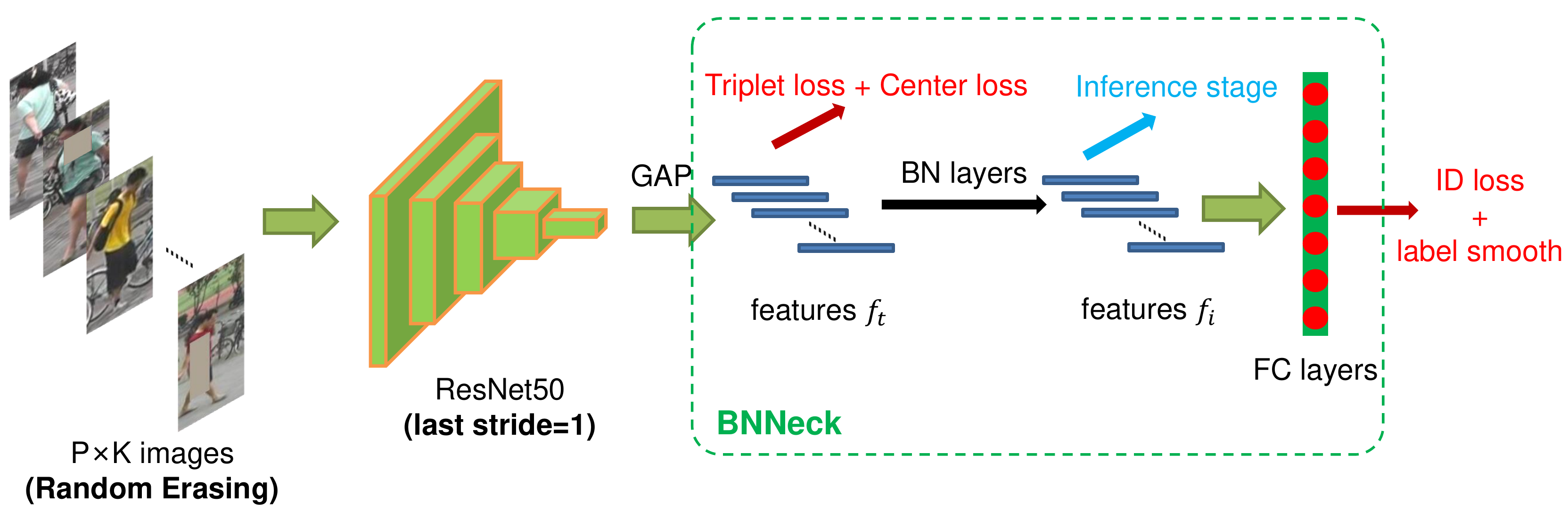}
\end{minipage}
\label{fig:arc_second}}
\centering
\caption{Pipelines of the standard baseline and our modified baseline.}
\label{fig:arc}
\end{figure*}

\section{Our Strong Baseline and Training Tricks}
A widely used baseline for the academia and industry is present in Fig. \ref{fig:arc_fisrt}.
For convenience, such baseline is called standard baseline.
Since we have introduced it in \cite{luo2019bags}, the details of such standard baseline can be found in \cite{luo2019bags} and our open source code.

In addition, this section introduces some effective training tricks in person ReID. Our proposed BNNeck structure is discussed in detail. The intra-class compactness and inter-class separability problem for person ReID is also raised. Most tricks can be expanded on the standard baseline without changing the model architecture.
Compared with our previous papaer \cite{luo2019bags}, this paper does not add any new contents for the first four tricks.
So we will introduce them simply and mainly focus on our proposed BNNeck and center loss.
Fig. \ref{fig:arc_second} shows training strategies and model architecture.

\subsection{Warmup Learning Rate}


Learning rate has a great effect on the performance of a ReID model. Standard baseline is initially trained with a large and constant learning rate.
In \cite{fan2019spherereid}, a warmup strategy was applied to bootstrap the network for enhanced performance.
In practice, we spend 10 epochs, thereby linearly increasing the learning rate from $3.5\times 10^{-5}$ to $3.5\times 10^{-4}$.
The learning rate is decayed to $3.5\times 10^{-5}$ and $3.5\times 10^{-6}$ at 40th and 70th epochs, respectively.
The learning rate $lr(t)$ at epoch $t$ is compute as follows:
\begin{equation}
lr(t)=\left\{\begin{array}{ll}
{3.5 \times 10^{-4} \times \frac{t}{10} }  & {\textnormal{ if } t \leq 10} \\
{3.5 \times 10^{-4}} & {\textnormal{ if } 10< t \leq 40} \\
{3.5 \times 10^{-5}}  & {\textnormal{ if } 40< t \leq 70} \\
{3.5 \times 10^{-6}} & {\textnormal{ if } 70< t \leq 120} \\
\end{array}\right.
\end{equation}

\subsection{Random Erasing Augmentation}
\begin{figure}[htb]
\centering
\includegraphics[width=.8\linewidth]{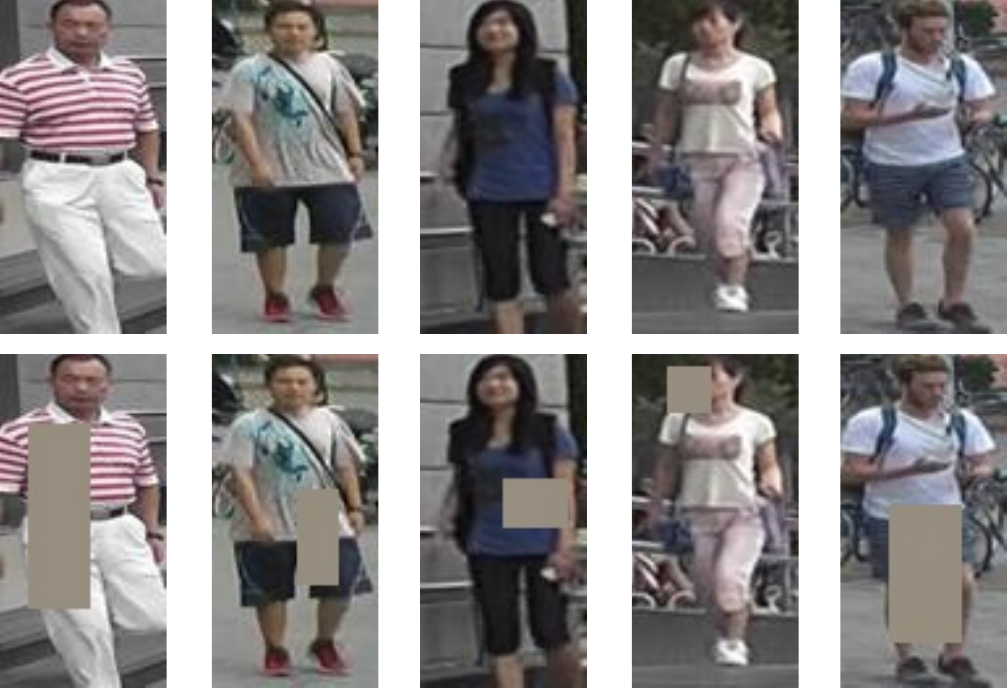}
\caption{Examples of random erasing augmentation. The first row shows the five original training images. The second row presents the processed images.}
\label{fig:rea}
\vspace{-8mm}
\end{figure}

In person ReID, persons in the images are sometimes occluded by other objects.
To address the occlusion problem and improve the generalization capability of ReID models, Zhong \emph{et al.} \cite{zhong2017random} proposed a new data augmentation approach, namely, random erasing augmentation (REA).
REA randomly masks a rectangle region of the training image with a manually set probability $p$.
All hyper-parameters are set same with \cite{zhong2017random} and \cite{luo2019bags}.
%
Some examples are shown in Fig. \ref{fig:rea}.

\subsection{Label Smoothing}
A basic baseline in person ReID is the IDE \cite{zheng2018discriminatively} network, whose last layer outputs the ID prediction logits of images.
Given an image, we denote $y$ as truth ID label and $p_i$ as ID prediction logits of class $i$.
The ID loss is computed as follows:
\begin{equation}
L(ID)= \sum_{i=1}^{N}-q_{i} \log \left(p_i\right) \left\{\begin{array}{ll}
q_i = 0, y \ne i \\
q_i = 1, y=i
\end{array}\right.
\end{equation}

However, in person ReID, person IDs of the testing set do not appear in the training set.
So it is important to prevent from overfitting training IDs for the ReID model.
A widely used technique to prevent overfitting for a classification task is Label smoothing (LS) proposed in \cite{szegedy2016rethinking}.
The construction of $q_i$ is changed to:
\begin{equation}\label{LS}
  q_{i}=\left\{\begin{array}{ll}
  {1- \frac{N-1}{N} \varepsilon} & {\textnormal{ if } i=y} \\
  {\varepsilon /N} & {\textnormal{ otherwise, }}\end{array}\right.
\end{equation}
where $\varepsilon$ is a constant to encourage the model to be less confident on the training set.
In this study, $\varepsilon$ is set to be $0.1$.
When the training set is not large, LS can significantly improve the model performance.

\subsection{Last Stride}
A high spatial resolution always enriches feature granularity.
In PCB\cite{sun2018beyond}, the last spatial down-sampling operation of the backbone network is removed to enlarge the spatial size of the feature map.
For convenience, the last spatial down-sampling operation in the backbone network is denoted as the last stride.
The last stride equals to 2 for ResNet50 backbone. When fed into an image with $256 \times 128$ size, it outputs a feature map with a spatial size of $8 \times 4$.
If last stride is changed from 2 to 1, then we can obtain a feature map with increased spatial size ($16 \times 8$).
This manipulation only slightly increases the computation cost and does not involve extra training parameters.
However, an increased spatial resolution brings significant improvement.

\subsection{BNNeck}

\begin{figure}[htb]
\vspace{-3mm}
\centering
\subfloat[Neck of the standard baseline.]{
\begin{minipage}[b]{0.25\textwidth}
\includegraphics[width=1\textwidth]{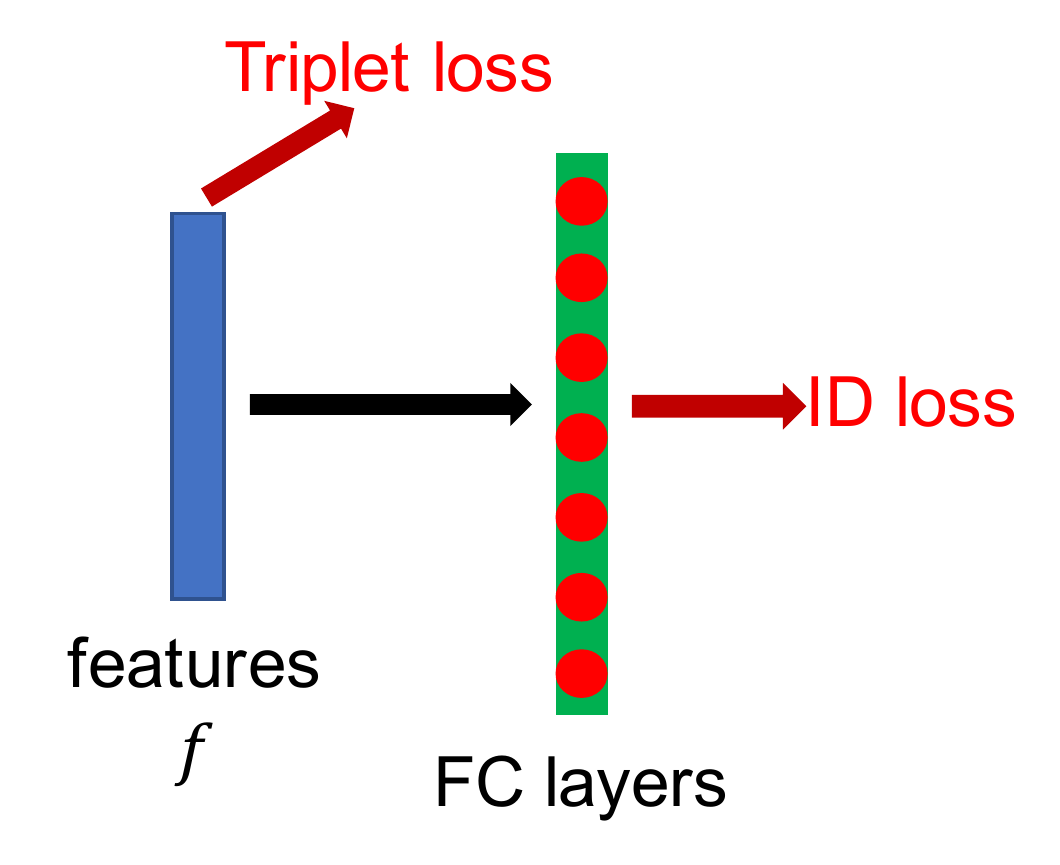}
\end{minipage}
}

\vspace{-3mm}
\centering
\subfloat[Designed BNNeck. In the inference stage, we select $f_i$ following the BN layer to perform the retrieval.]{
\begin{minipage}[b]{0.35\textwidth}
\includegraphics[width=1\textwidth]{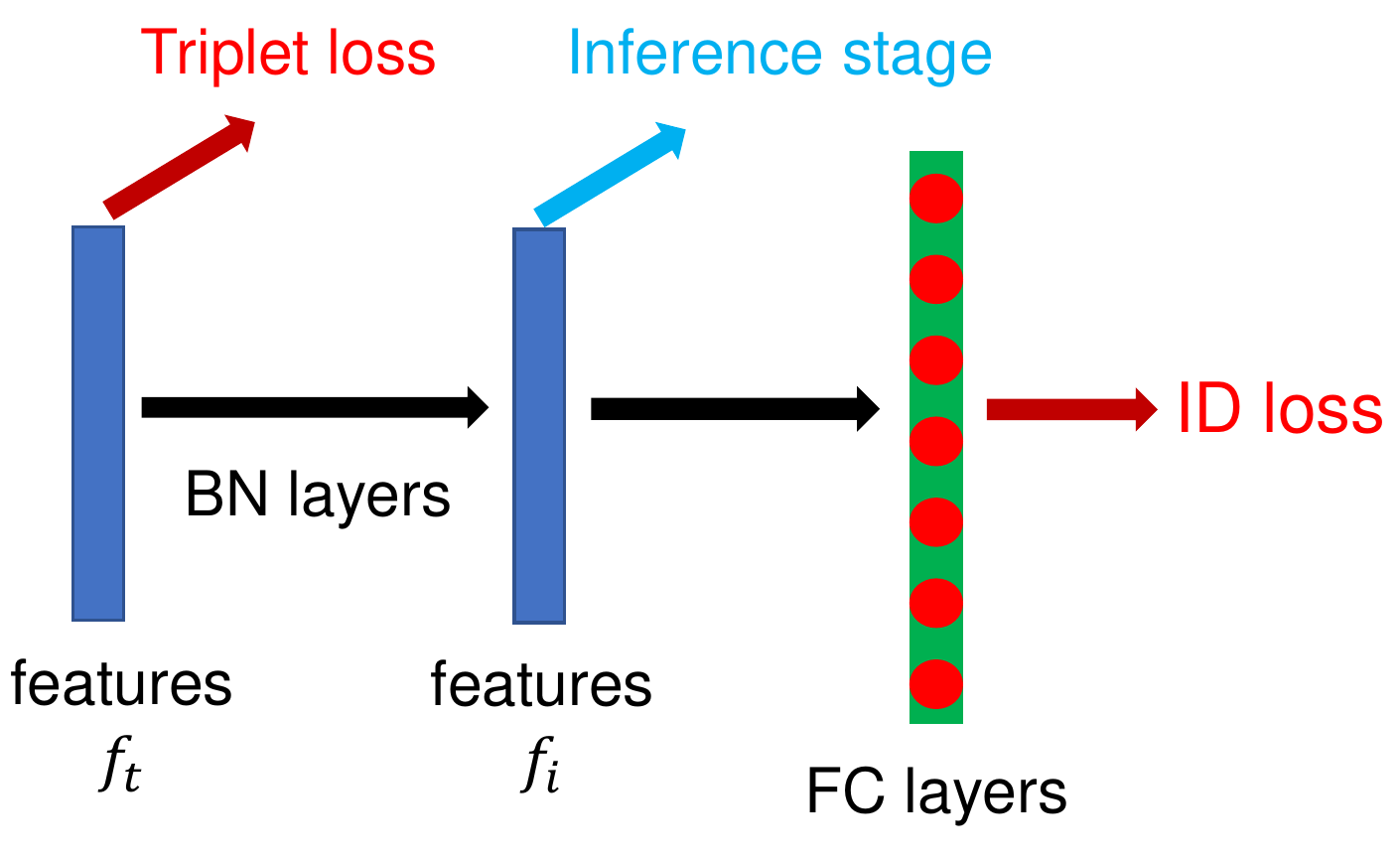}
\end{minipage}
}\caption{Comparison between standard neck and our designed BNNeck.}\label{fig:bnneck}
\end{figure}

\begin{figure*}[htb]
\centering
\vspace{-3mm}
\includegraphics[width=.99\linewidth]{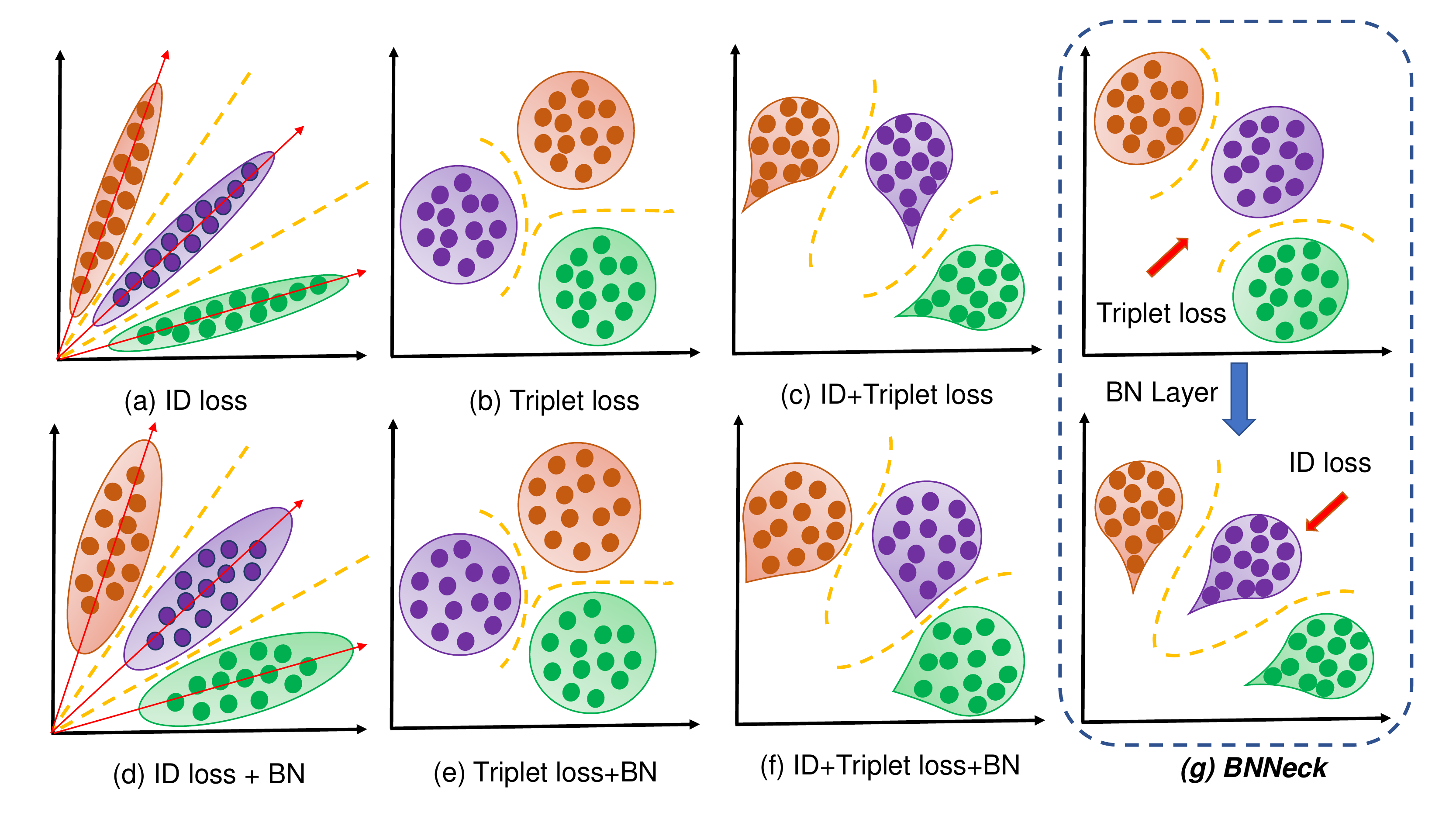}
\vspace{-7mm}
\caption{Two-dimensional visualization of sample distribution in the embedding space supervised by different losses and neck structures. (a$\sim$g) correspond to (a$\sim$g) in Fig. \ref{fig:necks}.
Points of different colors represent embedding features from different person IDs. The yellow dotted lines stand for decision surfaces. For better understanding, we make some overexpression compared to Fig. \ref{fig:vis}.}
\label{fig:eb}
\end{figure*}

As shown in Fig. \ref{fig:bnneck}(a), many state-of-the-art methods combined ID and triplet losses to constrain the same feature $f$.
Combining these two losses always let the model achieve better performance.
However, the better performance let us ignore that the targets of these two losses are inconsistent in the embedding space.

Fig. \ref{fig:eb}(a) presents that ID loss constructs several hyperplanes to separate the embedding space into different subspaces.
The features of each class are distributed affinely in different subspaces.
So ,cosine distance is more suitable than Euclidean distance for the model optimized by ID loss in the inference stage.
However, triplet loss is computed by Euclidean distance and enhances intra-class compactness and inter-class separability in the Euclidean space.
As shown in \ref{fig:eb}(b), The features of triplet loss appear a cluster distribution.
If we use both losses to optimize a feature space simultaneously, then their goals may be inconsistent.
During training, a possible problem is that one loss is reduced, whereas the other loss oscillates or even increases, as shown in Fig.\ref{fig:loss}.
Finally, triplet loss may influence the clear decision surfaces of ID loss, and ID loss may reduce the intra-class compactness of triplet loss.
The feature distribution is tadpole shaped. Therefore, directly combining these two losses can boost the performance, but it is not the best way.

Xiong \emph{et al.} \cite{xiong2019good} added a BN layer \cite{ioffe2015batch} iongbetween feature and ID loss, which is same as Fig. \ref{fig:necks}(d).
The authors claimed that the BN layer overcomes the overfitting and boosts the performance of IDE baseline.
However, we consider that the BN layer can smoothen the feature distribution in the embedding space.
For ID loss (Fig. \ref{fig:eb}(a)), the BN layer will enhance the intra-class compactness.
The BN layer can improve the performance of ID loss because the features close to the affine center lack clear decision surfaces and are difficult to distinguish.
Nevertheless, such layer increases the cluster radius of intra-class feature for triplet loss.
Thus, the decision surfaces of \ref{fig:eb}(e)(f) are stricter than those of Fig. \ref{fig:eb}(b)(c).

To overcome this problem, we design a structure, namely, BNNeck, as shown in Fig. \ref{fig:bnneck}(b).
BNNeck adds a BN layer after features and before classifier FC layers. The BN and FC layers are initialized through Kaiming initialization proposed in \cite{he2015delving}.
The feature before the BN layer is denoted as $f_t$. We let $f_t$ pass through a BN layer to acquire the feature $f_i$. In the training stage, $f_t$ and $f_i$ are used to compute triplet and ID losses, respectively.
Fig. \ref{fig:bnneck}(g) shows that $f_t$ not only can keep a compact distribution from but also acquires ID knowledge from ID loss.
Affected by the BN layer and ID loss, the distribution of $f_i$ is tadpole shaped.
In comparison with \ref{fig:bnneck}(c), $f_i$ has clear decision surfaces because of the weaker influence of the triple loss.
Additional details are introduced in Section \ref{sec:bnneck}.

In the inference stage, we select $f_i$ to perform the person ReID task. Cosine distance metric can achieve better performance than Euclidean distance metric. Experimental results in Table. \ref{ablation} show that BNNeck can improve the performance of the ReID model by a large margin.

\subsection{Center Loss}
\begin{figure}[htb]
\centering
\vspace{-3mm}
\includegraphics[width=.99\linewidth]{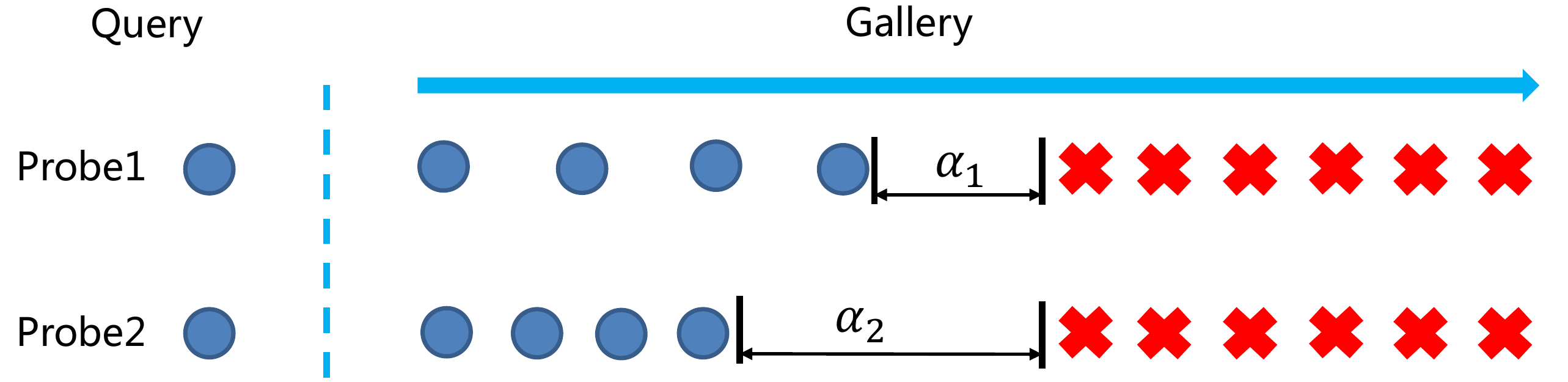}
\vspace{-7mm}
\caption{Visualized demonstration that ranking task ignores the intra-class compactness of positive pairs. Blue circles and red crosses represent positive and negative samples, respectively. In the direction of the arrow, the feature distance of two samples is increasing. Although two cases can acquire the same ranking results, probe2 is easy for the tracking task that must decide on a threshold to separate positive and negative samples.}
\label{fig:center}
\end{figure}

Person ReID is always regarded as a retrieval/ranking task.
The evaluation protocols, \emph{i.e.} CMC curve and mAP, are determined by the ranking results but ignore the clustering effect.
However, for some ReID applications, such as tracking task, an important step is to decide on a distance threshold to separate positive and negative objects. As shown in Fig. \ref{fig:center}, two cases can acquire the same ranking results, but probe2 is easy for the tracking task because of its intra-class compactness of positive pairs.

Focusing on relative distance, triplet loss is computed as:
\begin{equation}\label{eq:tri}
  L_{Tri} = [d_p - d_n + \alpha]_{+},
\end{equation}
where $d_p$ and $d_n$ are feature distances of positive and negative pairs.
$\alpha$ is the margin of triplet loss, and $[z]_{+}$ equals $max(z,0)$.
In this study, $\alpha$ is set to $0.3$.
However, the triplet loss only considers the difference between $d_p$ and $d_n$ and ignores their absolute values.
For instance, when $d_p=0.3$ and $d_n=0.5$, the triplet loss is $0.1$.
For another case, when $d_p=1.3$ and $d_n=1.5$, the triplet loss also is $0.1$.
Triplet loss is determined by two randomly sampled person IDs. Ensuring that $d_p < d_n$ in the entire training dataset is difficult. In addition, intra-class compactness is ignored.

To compensate for the drawbacks of the triplet loss, we involve center loss \cite{wen2016discriminative} intraining, simultaneously learns a center for deep features of each class and penalizes the distances between the deep features and their corresponding class centers. The center loss function is formulated as follows:
\begin{equation}\label{center}
  \mathcal{L}_{C}=\frac{1}{2} \sum_{j=1}^{B}\left\|\boldsymbol{f}_{t_j}-\boldsymbol{c}_{y_{j}}\right\|_{2}^{2},
\end{equation}
where $y_{j}$ is the label of the $j$th image in a mini-batch.
$\boldsymbol{c}_{y_{j}}$ denotes the $y_i$th class center of deep features, and $B$ is the batch size number. The formulation effectively characterizes the intra-class variations. Minimizing center loss increases intra-class compactness. Our model includes three losses as follows:
\begin{equation}\label{loss}
  L = L_{ID} + L_{Triplet} + \beta L_{C}
\end{equation}
where $\beta$ is the balanced weight of center loss.
In our baseline, $\beta$ is set to be $0.0005$.

\section{Experiment}

\subsection{Datasets}
We evaluate our models on Market1501 \cite{zheng2015scalable} and DukeMTMC-reID \cite{ristani2016MTMC} datasets, because both datasets are widely used and large scale.
Following the previous works, we use rank-1 accuracy and mAP for evaluation on both datasets.

\textbf{Market1501} contains 32,217 images of 1,501 labeled persons of six camera views. The training set has 12,936 images from 751 identities, and the testing set has 19,732 images from 750 identities. In testing, 3,368 hand-drawn images from 750 identities are used as queries to retrieve the matching persons in the database. Single-query evaluation is used in this study.

\textbf{DukeMTMC-reID}  is a new large-scale person ReID dataset and collects 36,411 images from 1,404 identities of eight camera views. The training set has 16,522 images from 702 identities, and the testing set has 19,889 images from other 702 identities. Single-query evaluation is used in this study.

\subsection{Influences of Each Trick}

\renewcommand{\multirowsetup}{\centering}
\vspace{-3mm}
\begin{table}[htb]\small
  \begin{center}
  \begin{tabular}{ l|cc|cc}
\hline
    			& \multicolumn{2}{c|}{Market1501} & \multicolumn{2}{c}{DukeMTMC}	 \\
  Model			& r = 1 	& mAP	&r = 1 	& mAP 	 \\
 	\hline
	\hline
Baseline-S		&87.7	&74.0	&79.7	&63.7			\\
+warmup         &88.7	&75.2	&80.6	&65.1           \\
+REA            &91.3	&79.3	&81.5	&68.3			\\
+LS	            &91.4	&80.3	&82.4	&69.3			\\
+stride=1	    &92.0	&81.7	&82.6	&70.6			\\
+BNNeck	        &94.1	&85.7	&86.2	&75.9		\\
+center loss    &94.5	&85.9	&86.4	&76.4		\\
\hline
  \end{tabular}
  \end{center}
  \caption{\label{ablation}Performance of different models is evaluated on Market1501 and DukeMTMC-reID datasets. Baseline-S represents the standard baseline.}
\vspace{-3mm}
\end{table}

\renewcommand{\multirowsetup}{\centering}
\begin{table}[htb]\small
  \begin{center}
  \begin{tabular}{ l|cc|cc}
\hline
    			& \multicolumn{2}{c|}{M$\to$D} & \multicolumn{2}{c}{D$\to$M}	 \\
  Model			& r = 1 	& mAP	&r = 1 	& mAP 	 \\
 	\hline
	\hline
Baseline		&24.4	&12.9	&34.2	&14.5			\\
+warmup         &26.3	&14.1	&39.7	&17.4           \\
+REA            &21.5	&10.2	&32.5	&13.5			\\
+LS	            &23.2	&11.3	&36.5	&14.9			\\
+stride=1	    &23.1	&11.8	&37.1	&15.4			\\
+BNNeck	        &26.7	&15.2	&47.7	&21.6			\\
+center loss	&27.5	&15.0	&47.4	&21.4			\\
-REA            &41.4	&25.7	&54.3	&25.5			\\
\hline

  \end{tabular}
  \end{center}
  \caption{\label{cd}The performance of different models is evaluated on cross-domain datasets. M$\to$D means that we train the model on Market1501 and evaluate it on DukeMTMC-reID.}
\end{table}

\begin{figure*}[tb]
\centering
\includegraphics[width=1.0\linewidth]{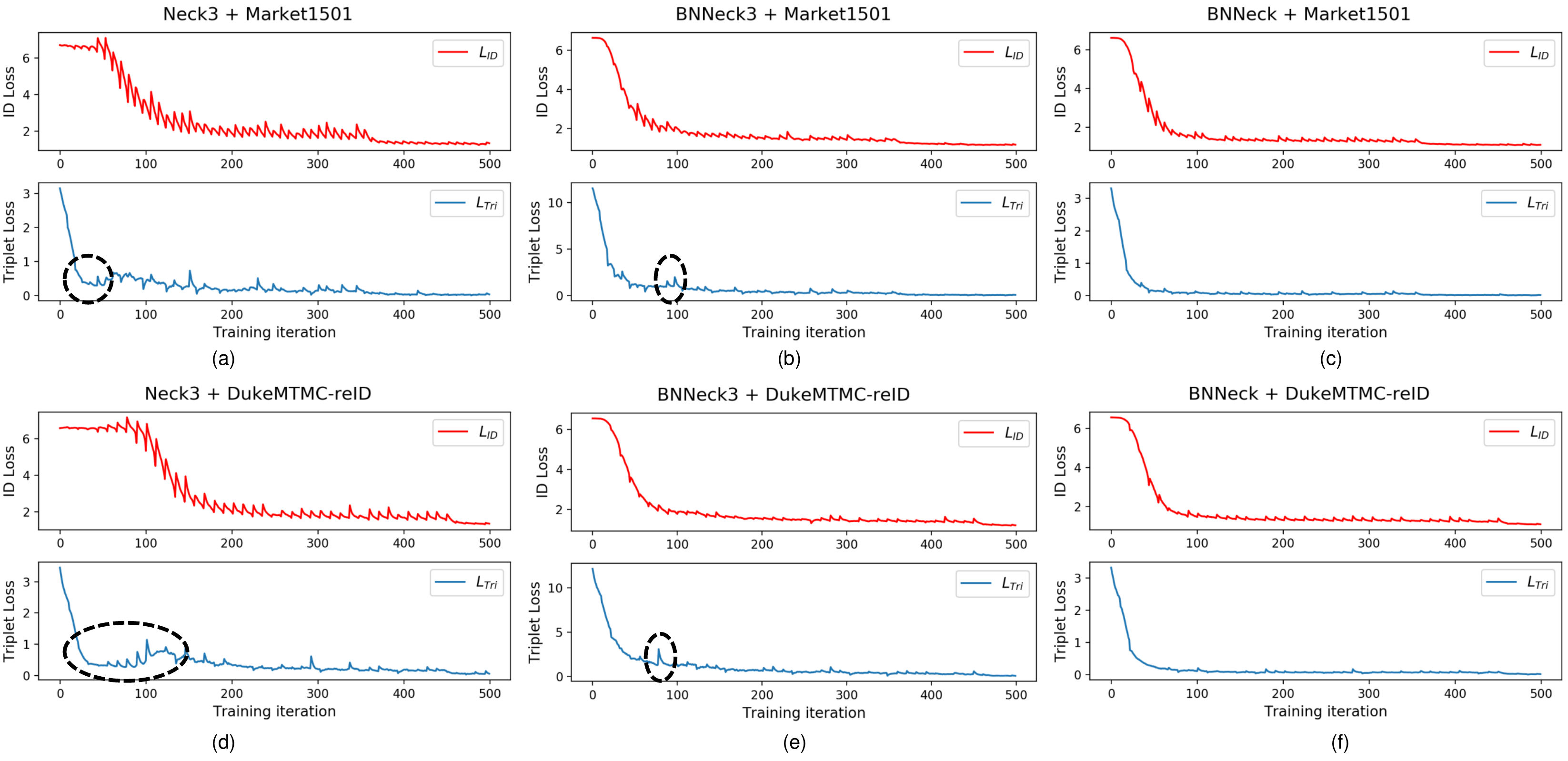}
\vspace{-9mm}
\caption{ID and triplet loss curves of different models on Market1501 and DukeMTMC-reID datasets. We train the models with Neck3, BNNeck3, and our proposed BNNeck, respectively. Black ovals mark the inconsistency between ID and triplet losses. We show that BNNeck suppresses the inconsistency and smoothens the triplet loss curve.}
\label{fig:loss}
\vspace{-5mm}
\end{figure*}

Baseline-S only reaches 87.7\% and 79.7\% rank-1 accuracies on Market1501 and DukeMTMC-reID, respectively.
The performance of standard baseline is similar to most baselines reported in other papers.
Warmup strategy, REA, LS, stride change, BNNeck, and center loss are individually added to the model training process.
The designed BNNeck boosts performance to a greater extent than other tricks, especially on DukeMTMC-reID.
Finally, the baseline acquires 94.5\% rank-1 accuracy and 85.9\% mAP on Market1501 with these training tricks.
On DukeMTMC-reID, the baseline achieves 86.4\% rank-1 and 76.4\% mAP accuracy .
Thus, these training tricks boost the performance of the standard baseline by over 10\% mAP.
In order to achieve such improvement, our strong baseline only involves an extra BN layer and do not increase training time.

We also explore the the effectiveness of these tricks for cross domain ReID, the results can be present in \ref{cd}.
In overall, most of these tricks apart from REA are also effective for cross domain ReID.
We infer that by REA masking the regions of training images, the model learns additional knowledge in the source domain and performs poorly in the target domain.

\subsection{Analysis of BNNeck}\label{sec:bnneck}

\subsubsection{Different neck structures}
\begin{figure}[tb]
\centering
\includegraphics[width=1.0\linewidth]{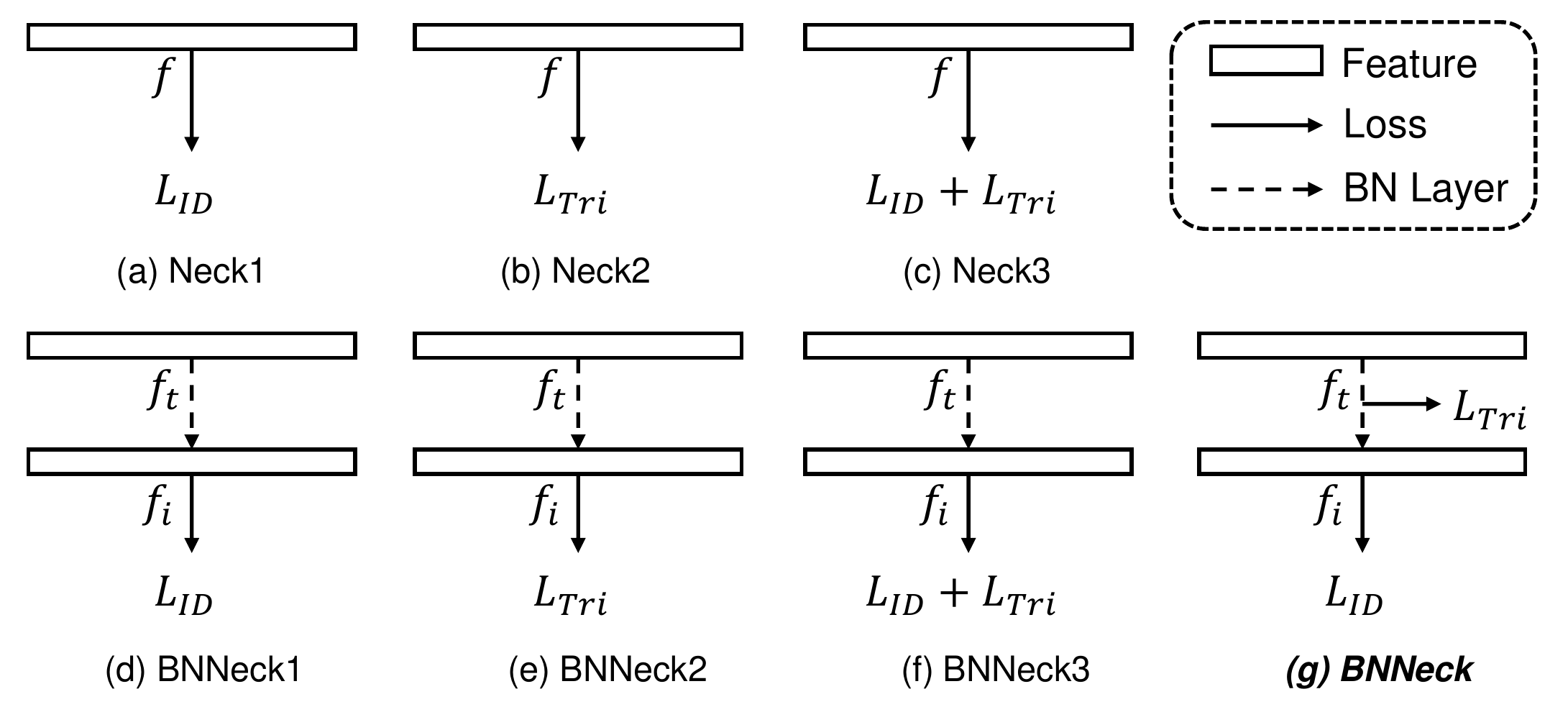}
\caption{Different neck structures for ablation study. (a$\sim$c) are standard neck structures, and (d$\sim$f) add an additional BN layer. Different losses includes $L_{ID}$, $L_{Tri}$, and $L_{ID}+L_{Tri}$.
(g) is our proposed BNNeck that separates triplet and ID losses into different feature spaces.}
\label{fig:necks}
\end{figure}

To discuss the effectiveness of our BNNeck, we design several different neck structures, as shown as Fig. \ref{fig:necks}.
In addition, some ablation studies also are analysed in Table \ref{tab:necks}.
Neck3 outperforms Neck1 and Neck2. In addition, BNNeck2 is worse than Neck2, but BNNeck1 is better than Neck1. Our BNNeck achieves the best performance on two benchmarks. In summary, we present the following observations/conclusions.
1) Without the BN layer, integrating ID and triplet losses is better than only using one loss.
2) The BN layer is effective for ID loss but is invalid for triplet loss.
3) Our BNNeck that sets triplet loss before the BN layer is a reasonable neck structure.

\begin{figure*}[htb]
\vspace{-2mm}
\centering
\includegraphics[width=.99\linewidth]{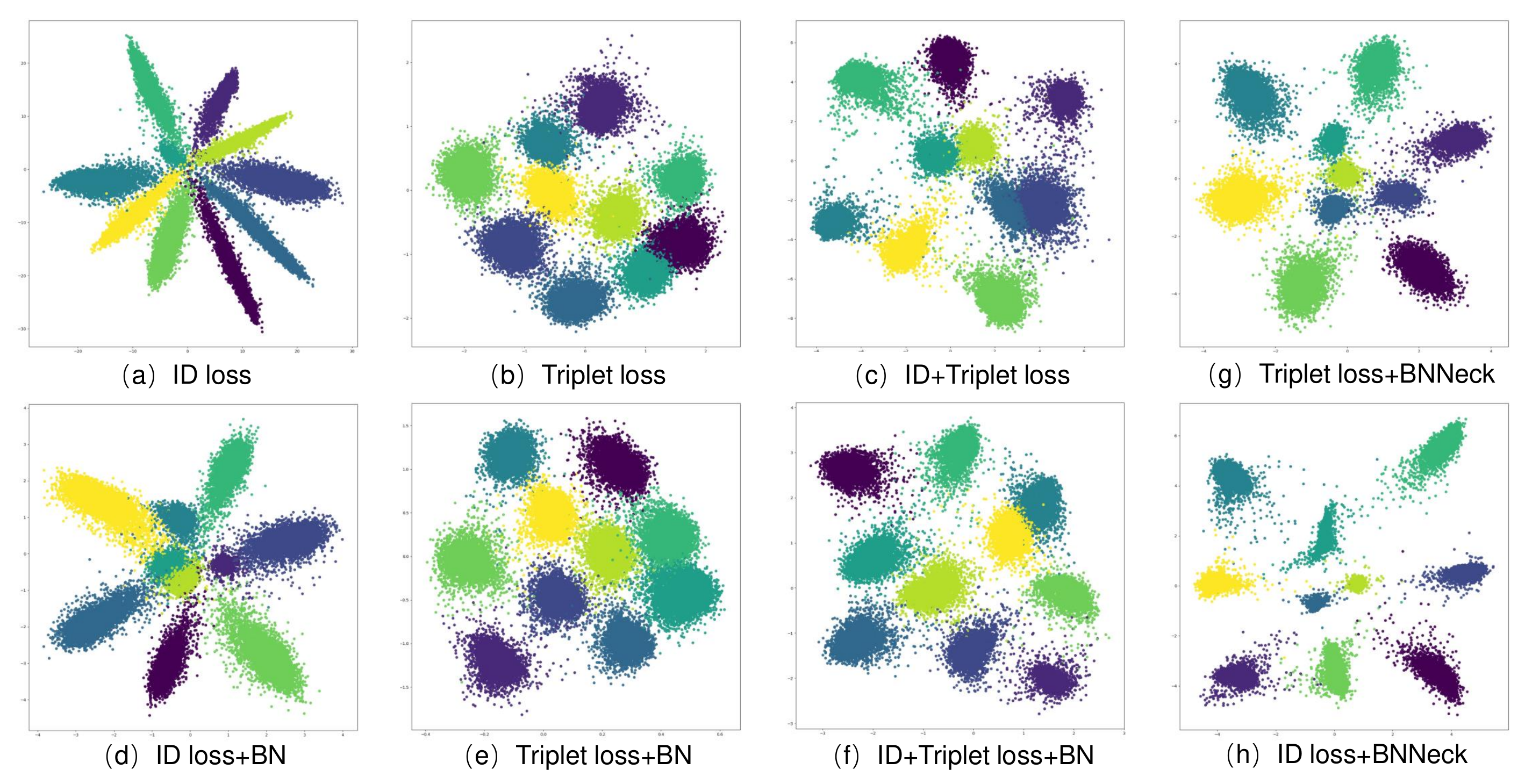}
\vspace{-3mm}
\caption{2D visualization of feature distribution in the embedding space supervised by different losses and neck structures on MNIST dataset. (a$\sim$f) correspond to (a$\sim$f) in Figs. \ref{fig:necks}. (g) and (h) are related to Fig. \ref{fig:necks}(g). The feature dimension is set to 2 for the best view. The BN layer will smoothen the feature.}
\label{fig:vis}
\vspace{-2mm}
\end{figure*}

\begin{figure*}
\centering
\subfloat[$f_t$ (before BN)]{
\begin{minipage}[b]{0.44\textwidth}
\includegraphics[width=1\textwidth]{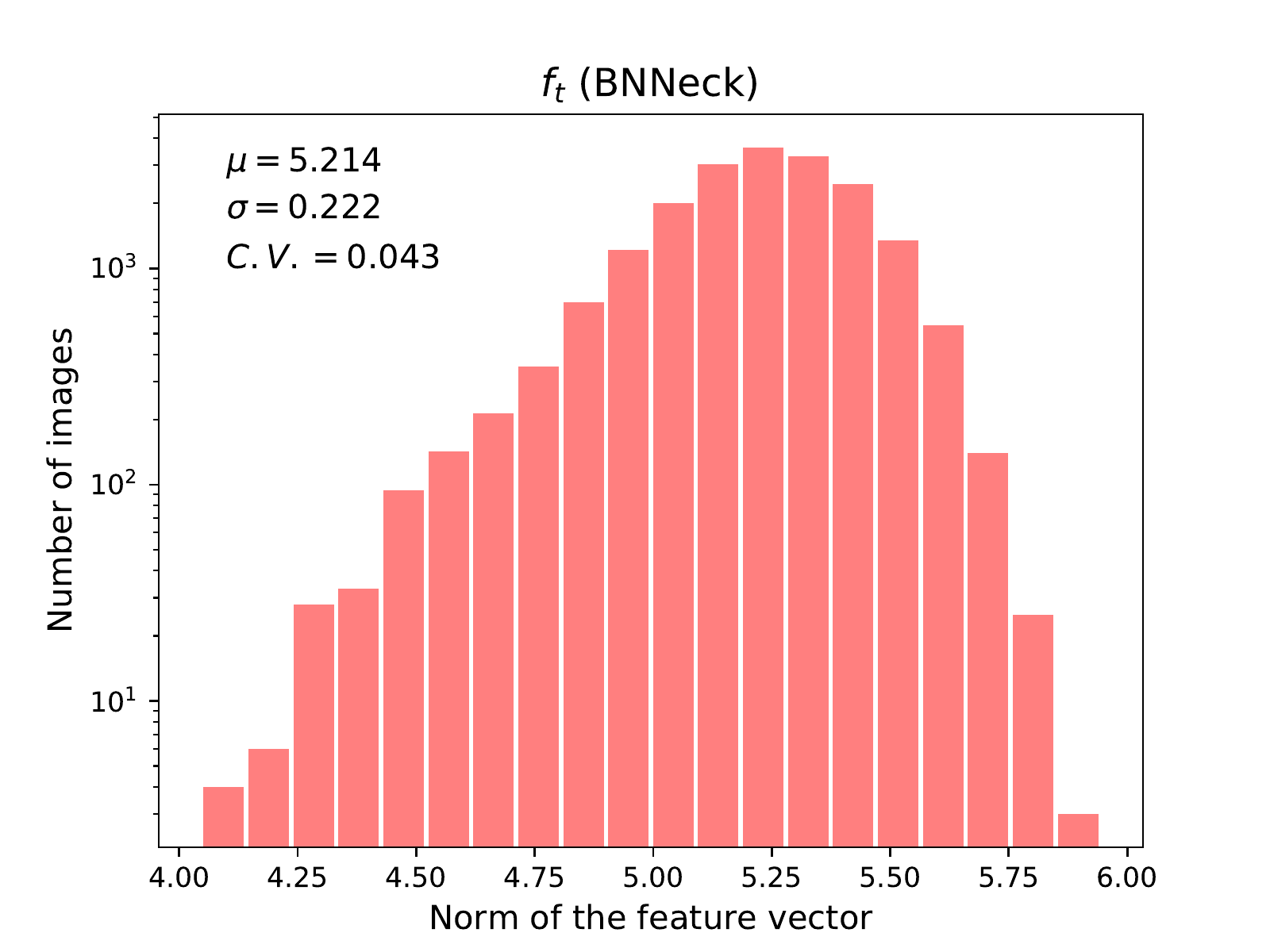}
\end{minipage}
}
\centering
\subfloat[$f_i$ (after BN)]{
\begin{minipage}[b]{0.44\textwidth}
\includegraphics[width=1\textwidth]{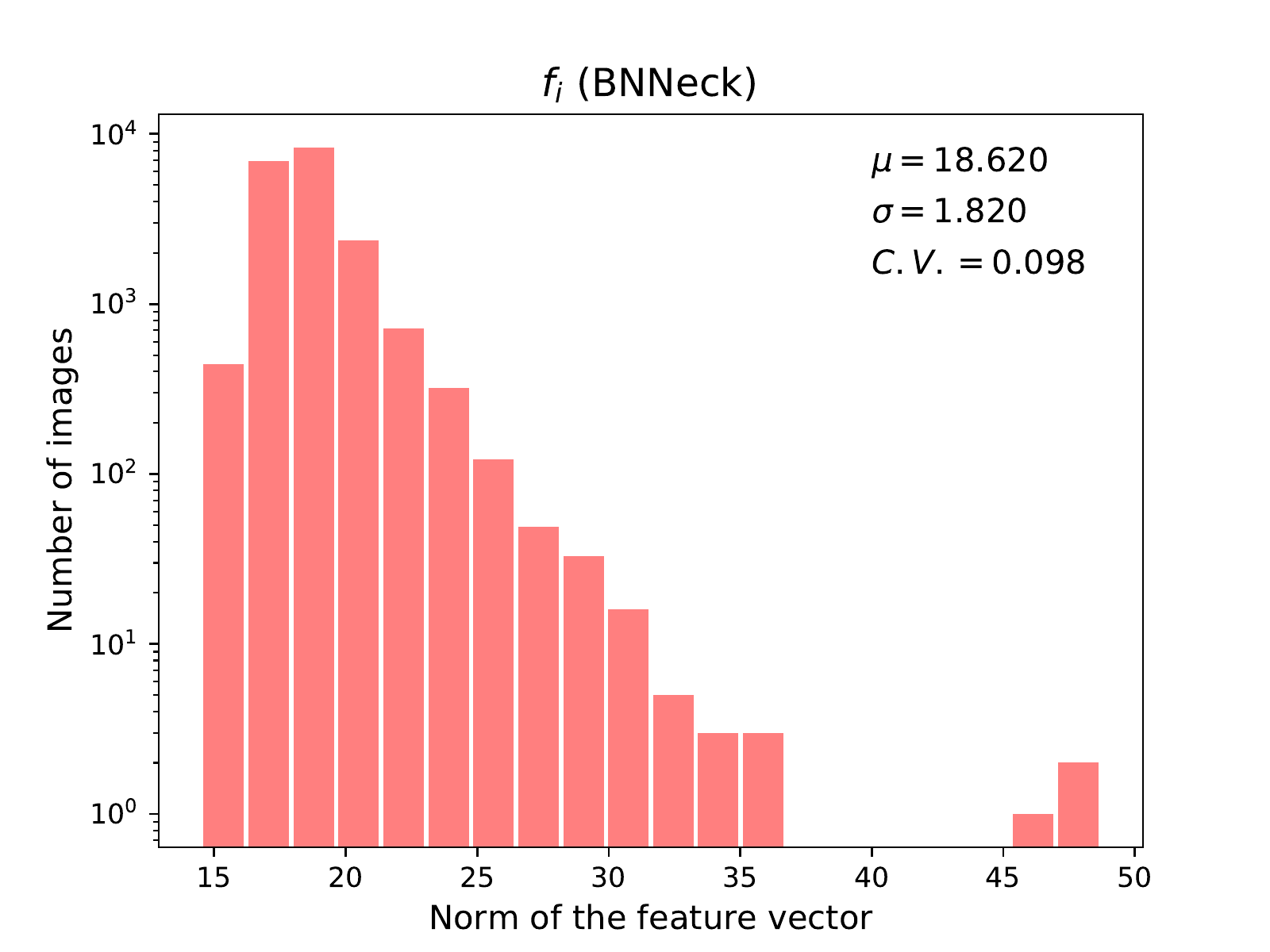}
\end{minipage}
}\caption{Histograms of feature norm for $f_t$ and $f_i$ in BNNeck on Market1501. $\mu$, $\sigma$, $C.V.$ are mean value, standard deviation, and Coefficient of Variation.}\label{fig:feat}
\vspace{-3mm}
\end{figure*}

\subsubsection{Inconsistency between ID loss and Triplet loss}

To verify that ID and triplet losses are inconsistent in the same feature space, we train the models with Neck3, BNNeck3, and our proposed BNNeck.
Fig. \ref{fig:necks} shows that these three neck structures use ID and triplet losses to optimize the same feature.
Fig. \ref{fig:loss} presents the training loss curves of 500 iterations.
In Fig. \ref{fig:loss}a and \ref{fig:loss}d, the triplet loss initially increases and then decays in the loss curves marked by black ovals, showing a clear confrontation between triplet and ID losses.
In comparison with Neck3, BNNeck3 adds a BN layer after f. In Figs. \ref{fig:loss}b and \ref{fig:loss}e, the BN layer weakens but does not eliminate the inconsistency.
However, for BNNeck in Figs. \ref{fig:loss}c and \ref{fig:loss}f, the inconsistency is suppressed, and the triplet loss curves are smoothened.
In conclusion, the BN layer can weaken the inconsistency between the losses, and separating them into two different feature spaces is important.

\subsubsection{Visualization of feature distribution}

\renewcommand{\multirowsetup}{\centering}
\begin{table}[tb]\small
  \begin{center}
  \begin{tabular}{ c|c|cc|cc}
\hline
    		&		       & \multicolumn{2}{c|}{Market1501} & \multicolumn{2}{c}{DukeMTMC}	 \\
  Feature   & Metric	   & r = 1 	& mAP  &r = 1 & mAP \\
 	\hline
	\hline
    $f$     & Neck1         &89.4	&77.5	&78.9	&65.3 \\
    $f$     & Neck2         &91.0	&80.9	&82.5	&69.4		\\
    $f$     & Neck3         &92.0	&81.7	&82.6	&70.6		\\
    $f_i$   & BNNeck1	    &93.1	&83.9	&85.2	&74.0       \\
    $f_i$   & BNNeck2	    &90.3	&79.1	&82.5	&67.9       \\
    $f_i$   & BNNeck3	    &92.5	&81.8	&83.3	&71.9       \\
    $f_t$   & BNNeck        &\textbf{94.2}&85.5 &85.7 &74.4		\\
    $f_i$   & BNNeck        &94.1	&\textbf{85.7} &\textbf{86.2}	&\textbf{75.9}		\\
\hline
  \end{tabular}
  \end{center}
  \caption{\label{tab:necks}Ablation study of different neck structures in Fig. \ref{fig:necks}.}
  \vspace{-2mm}
\end{table}

To analyze the distribution of the different features in Fig. \ref{fig:necks}, we train models in MNIST dataset. The visualization has considerable noise because the number of person IDs on ReID benchmark is large, and the number of images from each person ID is small. By contrast, MNIST only has 10 categories, and each category consists of thousands of samples, making the feature distribution clear and robust. Fig. \ref{fig:vis} shows the results. ID and triplet losses have two different feature distributions. When integrating these two losses in Fig. \ref{fig:vis}c, the clustered distribution is stretched to be tadpole shaped. The distributions of (d$sim$f) are more gaussian than those of (a$sim$c) because of the BN effect. Figs. \ref{fig:vis}g and \ref{fig:vis}h show that our BNNeck separates triplet and ID losses into two different feature spaces. The feature distribution of triplet loss remains clustered, and that of ID loss has clear decision surfaces similar to Figs. \ref{fig:vis}a and \ref{fig:vis}b.

We summarize our conclusions or observations as follows:
1) The feature distributions of ID and triplet losses are affined and clustered, \emph{i.e.}, they are inconsistent.
2) The feature distribution of ID+Triplet loss is tadpole shaped.
3) The BN layer can smoothen/normalize the feature distribution and enhance the intra-class compactness for ID loss but reduce it for triplet loss.
4) We separate triplet and ID losses into two different and suitable feature spaces.

\subsubsection{Two feature space of BNNeck}

Although the results on MNIST in Fig. \ref{fig:vis} can efficiently support our conclusion, image classification and person ReID are two different tasks. We perform statistical analysis on the norm distribution of $f_t$ and $f_i$ in BNNeck on Market1501 dataset.
The mean value $\mu$ and standard deviation $\sigma$ of feature norm are calculated.
To analyze the separability of feature distribution, Coefficient of Variation $C.V. = \mu/\sigma$ is also present.
As shown in Fig. \ref{fig:feat}, $f_i$ and $f_t$ are distributed differently in the feature space.
$f_t$ is compactly and gaussian distributed in an annular space because it is directly optimized by triplet loss.
However, we consider $f_i$ as a tadpole-shaped distribution because ID loss stretches intra-class distribution.
The maximum value of $f_i$ is 48.70, whereas the $\mu$ is 18.62.
$C.V.$ of $f_t$ is 0.043, but $C.V.$ of $f_i$ reaches 0.98, which demonstrates that $f_i$ is distributed more discretely than $f_t$.
In conclusion, BNNeck provide two different and suitable feature spaces for triplet loss and ID loss.

\subsubsection{Metric space for BNNeck}

We evaluate the performance of two different features ($f_t$ and $f_i$) with Euclidean and cosine distance metrics.
All models are trained without center loss in Table. \ref{tab:bnneck}.
We observe that cosine distance metric performs better than Euclidean distance metric for $f_t$.
As ID loss directly constrains the features followed the BN layer, $f_i$ can be clearly separated by several hyperplanes.
The cosine distance can measure the angle between feature vectors; thus, cosine distance metric is more suitable than Euclidean distance metric for $f_i$.
However, $f_t$ is simultaneously close to triplet loss and constrained by ID loss. The two types of metrics achieve similar performance for $f_t$.

Overall, BNNeck significantly improve the performance of ReID models.
We select $f_i$ with cosine distance metric to perform the retrieval in the inference stage.

\renewcommand{\multirowsetup}{\centering}
\begin{table}[tb]\small
  \begin{center}
  \begin{tabular}{ c|c|cc|cc}
\hline
    		&		       & \multicolumn{2}{c|}{Market1501} & \multicolumn{2}{c}{DukeMTMC}	 \\
  Feature   & Metric	   & r = 1 	& mAP	&r = 1 	& mAP 	 \\
 	\hline
	\hline
    $f$ (w/o BNNeck) & Euclidean    &92.0	&81.7	&82.6	&70.6 \\
    $f_t$   & Euclidean             &94.2	&85.5	&85.7	&74.4		\\
    $f_t$   & Cosine                &94.2	&85.7	&85.5	&74.6		\\
    $f_i$   & Euclidean	            &93.8	&83.7	&86.6	&73.0       \\
    $f_i$   & Cosine                &\textbf{94.1}	&\textbf{85.7}	&\textbf{86.2}	&\textbf{75.9}		\\
\hline
  \end{tabular}
  \end{center}
  \caption{\label{tab:bnneck}Ablation study of BNNeck. $f$ (w/o BNNeck) is baseline without BNNeck. BNNeck includes features $f_t$ and $f_i$. We evaluate their performance with Euclidean and cosine distance.}
\end{table}

\subsection{Analysis of Center loss}

\renewcommand{\multirowsetup}{\centering}
\begin{table}[tb]
  \begin{center}
  \begin{tabular}{ cl|ccc|ccc}
\hline
    	    &	& \multicolumn{3}{c|}{Market1501} & \multicolumn{3}{c}{DukeMTMC}	 \\
  Feature   & $\beta$		& r = 1  & mAP   & $R$   &r = 1  & mAP   & $R$ \\
 	\hline
	\hline
            &   0		    &94.2    &85.5   &0.407  &85.7   &74.4   &0.424\\
            &   0.0005		&93.9	 &85.7   &0.405	 &86.5	 &75.1   &0.420	\\
    $f_t$   &   0.005		&94.2	 &85.7   &0.394	 &86.2	 &75.4   &0.417	\\
            &   0.05		&94.4	 &85.4   &0.365	 &86.4	 &74.9   &0.403	\\
            &   0.5		    &92.6	 &81.1   &\textbf{0.311} &85.5	 &72.2   &\textbf{0.363}	\\
    \hline
            &   0		    &94.1    &85.7   &0.590  &86.2   &75.9   &0.568\\
            &   0.0005		&94.5	 &85.9   &0.589	 &86.4	 &76.4   &0.564	\\
    $f_i$   &   0.005		&94.3	 &85.9   &0.595	 &86.8	 &76.4   &0.566	\\
            &   0.05		&94.3	 &85.7   &0.592	 &86.7	 &76.5   &0.560	\\
            &   0.5		    &94.1	 &84.7   &0.593	 &87.4	 &76.9   &0.554	\\
\hline

  \end{tabular}
  \end{center}
  \caption{\label{tab:center} Evaluation with different weights of center loss $\beta$. $R$ is the ratio of intra-class distance to inter-class distance.}
\end{table}

We discuss the influence of center loss on intra-class compactness.
We consider that average intra-class distance cannot fully represent the intra-class compactness because it ignores inter-class distance.
For convenience, the average intra-class and inter-class distances are denoted as $D_p$ and $D_n$, respectively. Inspired by \cite{zhang2017range}, the ratio of $D_p$ to $D_n$ is used to measure the clustering effect of feature distribution. The ratio is computed as $R = D_p/D_n$. We set $\beta$ to different values and evaluate rank-1, mAP, and $R$ of the models. Table \ref{tab:center} presents the results.

For the feature $f_t$ constrained directly by center loss, $R$ decreases as $\beta$ increases.
With $\beta$ increasing from 0 to 0.5, $R$ is reduced from 0.407 to 0.311 on Market1501 and from 0.424 to 0.363 on DukeMTMC-reID.
Hence, center loss can improve intra-class compactness and inter-class separability, thereby bringing a clear boundary between positive and negative samples.
When $\beta$ is set to 0.5, $f_t$ can acquire the best clustering effect but obtains the worse rank-1 and mAP accuracies. However, the BN layer destroys such clustering effect.
For feature $f_i$, the value of $R$ is almost not influenced by $\beta$.
On the basis of these observations, we arrive at the following conclusions:
(1) Center loss boosts intra-class compactness and inter-class separability.
(2) The BN layer can destroy the effect of center loss.
(3) Increasing the weight of center loss may reduce ranking performance.

\subsection{Comparison to Other Baselines}

\renewcommand{\multirowsetup}{\centering}
\begin{table}[tb]\footnotesize
  \begin{center}
  \begin{tabular}{ c|c|cc|cc}
\hline
    		&		       & \multicolumn{2}{c|}{Market1501} & \multicolumn{2}{c}{DukeMTMC}	 \\
  Baseline   & Loss	       & r = 1 	& mAP	&r = 1 	& mAP 	 \\
 	\hline
	\hline
    IDE\cite{zheng2018discriminatively}     & ID            & 79.5	& 59.9	& -	&-            \\
    TriNet\cite{hermans2017defense}  & Tri           & 84.9	& 69.1	& -	& -		\\
    AWTL\cite{ristani2018features}    & Tri           & 89.5	& 75.7	& 79.8	& 63.4		\\
    GP\cite{xiong2019good}     & ID	        &91.7   &78.8	&83.4   &68.8       \\
    PCB\cite{sun2018beyond}     & ID            & 92.3	&77.4	&81.7	&66.9		\\
    Our     & ID+Tri        &\textbf{94.5}	&\textbf{85.9}	&\textbf{86.4}	&\textbf{76.4}		\\
\hline

  \end{tabular}
  \end{center}
  \caption{\label{tab:bs} Comparison of different baselines on Market1501 and DukeMTMC-reID datasets. ID and Tri stands for ID loss and triplet-based loss, respectively.}
\vspace{-2mm}
\end{table}

We compare our strong baseline with other effective baselines, such as IDE \cite{zheng2018discriminatively}, TriNet \cite{hermans2017defense}, AWTL \cite{ristani2018features} and PCB \cite{sun2018beyond}.
PCB is a part-based baseline for person ReID. Table \ref{tab:bs} presents the performance of these baselines. The experimental results show that our baseline outperforms IDE, TriNet, and AWTL by a large margin. PCB integrates multi-part features and GP uses effective tricks, and both of them achieves great performance. However, our baseline surpasses them by over 7.1\% mAP on both datasets. To our best knowledge, our baseline is the strongest baseline.

\subsection{Comparison to State-of-the-Arts}

\renewcommand{\multirowsetup}{\centering}
\begin{table}[tb]\footnotesize
  \begin{center}
  \begin{tabular}{ ccc|cc|cc}
\hline
    		&		&       & \multicolumn{2}{c|}{Market1501} & \multicolumn{2}{c}{DukeMTMC}	 \\
  Type   & Method & $N_f$	   & r = 1 	& mAP	&r = 1 	& mAP 	 \\
 	\hline
	\hline
    \multirow{3}{1cm}{Pose-guided}& GLAD\cite{wei2017glad}        & 4  &89.9	&73.9	&-	&-		\\
                                & PIE \cite{zheng2019pose}          & 3  &87.7	&69.0	&79.8	&62.0		\\
                                & PSE \cite{Sarfraz_2018_CVPR}      & 3  &78.7	&56.0	&-	&-		\\
    \hline
    \multirow{2}{1cm}{Mask-guided}& SPReID \cite{kalayeh2018human}    & 5  & 92.5 & 81.3	& 84.4	&71.0		\\
                                & MaskReID \cite{qi2018maskreid}    & 3  &90.0	&75.3	&78.8	&61.9		\\
 \hline
    \multirow{8}{1cm}{Stripe-based}& AlignedReID++ \cite{LUO2019}& 1 &90.6 &77.7 &81.2	&67.4 		\\
                                & SCPNet \cite{fan2018scpnet}       & 1  & 91.2	&75.2	&80.3	&62.6		\\
                                & PCB+RPP \cite{sun2018beyond}      & 6  & 93.8	&81.6	&83.3	&69.2		\\
                                & Pyramid\cite{zheng2018coarse}     & 1 & 92.8 &82.1	&-	&-		\\
                                & Pyramid\cite{zheng2018coarse}     & 21 & 95.7 &88.2	&89.0	&79.0		\\
                                & BFE\cite{dai2018batch}            & 2 & 94.5 &85.0	&88.7	&75.8		\\
                                & MGN \cite{wang2018learning}       & 1 & 89.8 &78.5	&-	&-		\\
                                & MGN \cite{wang2018learning}       & 8 & 95.7 &86.9	&88.7	&78.4		\\
   \hline
    \multirow{3}{1cm}{Attention-based}& Mancs \cite{wang2018mancs}        & 1 &93.1   &82.3	&84.9	&71.8		\\
                                & DuATM \cite{si2018dual}           & 1 & 91.4	& 76.6	&81.2	&62.3		\\
                                & HA-CNN \cite{li2018harmonious}    & 4 & 91.2	& 75.7	&80.5	&63.8		\\
    \hline
    \multirow{2}{1cm}{GAN-based}& Camstyle \cite{zhong2019camstyle} &  1 &88.1 	&68.7	&75.3	&53.5		\\
                                & PN-GAN \cite{qian2018pose}        &  9 &89.4 	&72.6	&73.6	&53.2		\\
    \hline
    \multirow{5}{1cm}{Global feature}& IDE \cite{zheng2018discriminatively}  & 1  & 79.5	& 59.9	& -	&-		\\
                                & SVDNet \cite{sun2017svdnet}  & 1  & 82.3	& 62.1	& 76.7	&56.8		\\
                                & TriNet\cite{hermans2017defense}  & 1  & 84.9	& 69.1	& -	& -		\\
                                & AWTL\cite{ristani2018features}  & 1  & 89.5	& 75.7	& 79.8	& 63.4		\\
                                & \textbf{Ours}    &  1 &\textbf{94.5}	&\textbf{85.9}	&\textbf{86.4}	&\textbf{76.4}		\\
\hline
  \end{tabular}
  \end{center}
  \caption{\label{tab:sota}Comparison of state-or-the-art methods. $N_f$ is the number of features used in the inference stage.}
\end{table}

We compare our strong baseline with some state-of-the-art methods in Table. \ref{tab:sota}.
All methods have been divided into different types.
Pyramid\cite{zheng2018coarse} achieves surprising performance on two datasets, but it concatenates 21 local features of different scales. When only the global feature is utilized, Pyramid obtains 92.8\% rank-1 accuracy and 82.1\% mAP on Market1501. Our strong baseline can reach 94.5\% rank-1 accuracy and 85.9\% mAP on Market1501.
BFE\cite{dai2018batch} obtains similar performance to our strong baseline, but it combines features of two branches.
Among all methods that only use global features, our strong baseline outperforms AWTL\cite{ristani2018features} by more than 10\% mAP on both Market1501 and DukeMTMC-reID.
To our best knowledge, our baseline achieves the best performance when only global features are used.
Our method, which does not require human semantic information, local features or attention modules, is simpler than other state-of-the-art methods.

\subsection{Baseline Meets State-of-the-Arts}

\renewcommand{\multirowsetup}{\centering}
\begin{table*}[tb]\footnotesize
  \begin{center}
  \begin{tabular}{ c|c|cc|cc|l}
\hline
    		&		       & \multicolumn{2}{c|}{Market1501} & \multicolumn{2}{c|}{DukeMTMC} &  	 \\
  Method    & Reference	   & r = 1 	& mAP	&r = 1 	& mAP &\multicolumn{1}{c}{Loss} 	 \\
 	\hline
	\hline
    Baseline1   &   BNNeck1 &93.1     &83.9	    &85.2        &74.0  & $L_{ID}$\\
    Baseline2   &   BNNeck  &94.1     &85.7	    &86.2        &75.9  & $L_{ID}$, $L_{Tri}$         \\
    Baseline3   &   BNNeck  &94.5	  &85.9	    &86.4	&76.4	  & $L_{ID}$, $L_{Tri}$, $L_{C}$ \\
    \hline
    $k$-reciprocal\cite{xiong2019good}  & CVPR17	        &95.4(77.1)     &94.2(63.6)	    &90.3(-)        &89.1(-)  & $L_{ID}$, $L_{Tri}$, $L_{C}$  \\
    PCB\cite{sun2018beyond}             & ECCV18            & 94.0(92.3)	& 84.0(77.4)	& 88.6(81.7)	& 77.2(66.1)  & $L_{ID}$, $L_{Tri}$		\\
    AligedReID++ \cite{LUO2019}     & PR19                  & 94.3(91.8)	& 86.5(79.1)	& 86.5(82.1)	&76.9(69.7)    & $L_{ID}$, $L_{Tri}$        \\
    CamStyle\cite{zhong2019camstyle}  & TIP19               & 93.3(88.1)	& 81.0(68.7)	& 80.3(75.3)	& 60.1(53.5)	 & $L_{ID}$	\\
    MGN\cite{wang2018learning}     & ACMMM19                & 95.3(95.7)	&86.3(86.9)	&89.2(88.7) &78.9(78.4)		 & $L_{ID}$, $L_{Tri}$\\
\hline

  \end{tabular}
  \end{center}
  \caption{\label{tab:boost} Performance of some state-of-the-art methods reproduced by our strong baseline. The values in parentheses are the results reported by authors.}
\vspace{-2mm}
\end{table*}
We reproduce some popular state-of-the-art methods with our strong baseline. Given numerous outstanding methods are available, we cannot try all of them and select only several typical models such as $k$-reciprocal re-ranking\cite{zhong2017re}, PCB\cite{sun2018beyond}, AligedReID++\cite{LUO2019}, CamStyle\cite{zhong2019camstyle}, and MGN\cite{wang2018learning}.
For a fair comparison, we use the same losses as the paper reported to train the models. 
For instance, AlignedReID++ only uses ID and triplet losses, and we do not use center loss to reproduce it. 
However, as $k$-reciprocal re-ranking is a post-processing method of global features, three losses are used to improve its performance. Table \ref{tab:boost} shows the details and results, wherein the values in parentheses are the results reported by authors in their papers. In addition, we present the performance of the baselines (with BNNeck) trained by different losses as a reference.

Our baseline boosts the performance of $k$-reciprocal re-ranking, PCB, AligedReID++, and CamStyle by a large margin. The mAP of k-reciprocal re-ranking achieves +30.6\% on Market1501, demonstrating that the performance of baselines is important for methods. In addition, our MGN achieves similar performance to \cite{wang2018learning} because its accuracies are too high to improve, and \cite{wang2018learning} uses BNNeck1 structure. 
Integrating multiple part features can reduce the effect of global features and limit the effectiveness of baselines for PCB and MGN. 
However, PCB and MGN still obtain better performance than Baseline2, \emph{i.e.}, part-based methods are effective for our baseline. 

However, CamStyle(Our) outperforms CamStyle \cite{zhong2019camstyle} but not Baseline1. 
Our baseline can be a strong baseline for the ReID community because it can boost the performance of some methods, and other methods based on it may be ineffective.
To some extent, our baseline efficiently filters effective methods.

\begin{table}[tb]\footnotesize
  \begin{center}
  \begin{tabular}{ c|cc|cc}
\hline
  	 				& \multicolumn{2}{c|}{Market1501} & \multicolumn{2}{c}{DukeMTMC}	 \\
  Backbone			& r = 1 	& mAP	&r = 1 	& mAP 	 \\
 	\hline
	\hline
ResNet18     &91.7	&77.8	&82.5	&68.8			\\
ResNet34	   &92.7 &82.7 &  86.4 &73.6 			\\
ResNet50     &94.5 &85.9 & 86.4 &76.4    		\\
ResNet101     &94.5 &87.1 &  87.6 &77.6			\\
SeResNet50    &94.4 &86.3 & 86.4 &76.5			\\
SeResNet101    &94.6 &87.3 & 87.5 &78.0			\\
SeResNeXt50    &94.9 &87.6 & 88.0 &78.3			\\
SeResNeXt101    & 95.0 &88.0 & 88.4 &79.0			\\
IBN-Net50-a    &\textbf{95.0}	&\textbf{88.2}	&\textbf{90.1}	&\textbf{79.1}			\\
IBN-Net50-b    &93.5	&83.9	&86.4	&73.5			\\
\hline
  \end{tabular}
  \end{center}
  \caption{\label{tab:backbone}Performance of our baseline with different backbone.}
\vspace{-2mm}
\end{table}

\subsection{Performance of Different Backbones}

All aforementioned models apply ResNet50 as backbones for clear ablation studies and comparison with other methods.

Models with different backbones, such as ResNet, SeResNet, SeResNeXt, and IBNNet, are evaluated because backbones have a great influence on their performance.
As shown in Table \ref{tab:backbone}, deep and large backbones can achieve high performance. For example, ResNet101 outperforms ResNet18 by 2.8\% and 9.3\% in Rank-1 and mAP accuracy on Market1501, respectively.
In addition, the channel attention of SeNet and group convolution of ResNeXt can enhance the performance by a slight margin.
IBN-Net50 \cite{pan2018two}, which replaces the BN layers with instance BN layers for ResNet50, is also effective for our baseline.
Specifically, IBN-Net50-a is suitable for standard ReID task and obtains 95.0\% and 90.1\% rank-1 accuracies on Market1501 and DukeMTMC-reID, respectively.
However, IBN-Net50-b achieves 50.1\% rank-1 and 29.8\% mAP for Market1501$\to$DukeMTMCReID (M$\to$D) and 61.7\% rank-1 and 32.0\% mAP DukeMTMCReID1501$\to$Market1501 (D$\to$M).

For comparison, IBN-Net50-a achieves 40.0\% rank-1 and 25.1\% mAP for M$\to$D and 52.9\% rank-1 and 25.1\% mAP (D$\to$M). In conclusion, IBN-Net-a and IBN-Net50-b are suitable for the same domain task and the cross-domain task, respectively.

\section{Conclusions and Outlooks}

In this study, we propose a strong baseline for person ReID with only adding an extra BN layer for standard baseline. Our strong baseline achieves 94.5\% rank-1 accuracy and 85.9\% mAP on Market1501. To our best knowledge, this result is the best performance achieved by the global features of a single backbone. We evaluate each trick of our baseline on same- and cross-domain ReID tasks. In addition, some state-of-the-art methods can be effectively extended on our baseline. We hope that this work can promote ReID research in the academia and industry.

We observe the inconsistency between ID and triplet losses in previous ReID baselines. To address this problem, we propose a BNNeck to separate both losses into two different feature spaces. Extended experiments show that the BN layer can enhance and reduce the intra-class compactness for ID and triplet losses, respectively. Furthermore, ID loss is suitable for optimizing the feature.

We emphasize that the evaluation of ReID task ignores the clustering effect of representation features. However, the clustering effect is important to some ReID applications, such as tracking task wherein an important step is deciding on a distance threshold to separate positive and negative objects. A simple way to address this problem is using center loss to train the model. Center loss can boost the clustering effect of features, but may reduce the ranking performance of ReID models.

In the future, we will explore additional tricks and effective methods based on this strong baseline. In comparison with face recognition, person ReID still has room for further exploration. In addition, some confusions remain, such as why REA reduces the cross-domain performance in our baseline. Points wherein the conclusion is unclear are worth researching.

\section*{Acknowledgment}

This research was funded by the National Natural Science Foundation of China under Grant 61633019, the Science Foundation of Chinese Aerospace Industry under Grant JCKY2018204B053 and the Autonomous Research Project of the State Key Laboratory of Industrial Control Technology, China (Grant No. ICT1917).

\ifCLASSOPTIONcaptionsoff
  \newpage
\fi

\bibliography{egbib}{}
\bibliographystyle{IEEEtran}



\end{document}